\documentclass[final,5p,times,twocolumn]{elsarticle}
\usepackage{amsmath}
\usepackage{amssymb} 

\usepackage{tikz}
\usetikzlibrary{arrows,matrix,positioning,patterns}
\usetikzlibrary{calc}
\usepackage{pgfplots} 
\usetikzlibrary{external}
\usetikzlibrary{angles}
\usepackage{hyperref}
\usetikzlibrary{plotmarks}

\newdefinition{rmk}{Remark}
\newproof{pf}{Proof}
\newproof{pot}{Proof of Theorem \ref{thm2}}

\usepackage[numbers]{natbib}
\usepackage[ruled, vlined, linesnumbered]{algorithm2e}

\usepackage{amsthm}

\newtheorem{problem}{Problem}

\usepackage{amsmath}
\usepackage{amssymb} 

\usepackage{eurosym}
\usepackage{colortbl}

\usetikzlibrary{arrows.meta,calc,decorations.markings,math,arrows.meta}

\usepackage[labelformat=simple]{subcaption}
\DeclareCaptionLabelSeparator{periodspace}{.\quad}
\usepackage{color,framed}
\usepackage[utf8]{inputenc} 
\usepackage{tabularx}

\usepackage{rotating}
\usepackage{graphicx}
\usepackage{lscape}
\allowdisplaybreaks 
\usepackage{longtable}

\usepackage[utf8]{inputenc}
\usepackage{pgfplots}
\usepgfplotslibrary{groupplots} 
\pgfplotsset{compat=1.3}

\makeatletter
\def\ps@pprintTitle{%
 \let\@oddhead\@empty
 \let\@evenhead\@empty
 \def\@oddfoot{}%
 \let\@evenfoot\@oddfoot}
\makeatother

\begin{document}

\begin{frontmatter}
%
%
\title{Distributed area coverage path planning subject to the inclusion of a headland path}
\author{Mogens Plessen\corref{cor1}}
\cortext[cor1]{MP is with Findklein GmbH, Switzerland, \texttt{pmogens@proton.me}}

%
%
%
%

\begin{abstract}
A deterministic method for the problem of distributed area coverage path planning subject to the inclusion of a headland path is presented. Given a, in general, non-convex area to be covered that may, in general, include obstacle areas, multiple robots, at least one robot depot, and constraints on each machinery such as a working width, work capacity limits, travel capacity limits, nonholonomic dynamical constraints and, possibly, a robot start and end depot assignment, distributed paths for each machinery are calculated that collectively enable full area coverage. Variations are discussed, in particular, a scenario where unconstrained depots assignment can attain a global optimal solution to the problem. The importance of inclusion of a headland path for area coverage, as well as the benefits in comparison to an alternative Boustrophedon-pattern based approach are highlighted. 
\end{abstract}
\end{frontmatter}


\section{Introduction\label{sec_intro}}

\begin{table}
\centering
\begin{tabular}{|ll|}
\hline
\multicolumn{2}{|c|}{MAIN NOMENCLATURE}\\
\multicolumn{2}{|l|}{Symbols}\\
$L$ & Pathlength, (m).\\
$D$ & Depot location, (-).\\
$N$ & Nr. of robots, (-).\\
$R_i$ & Turning radius of robot $i$, (m).\\
$T$ & Transition location, (-).\\
$W$ & Working width (operating width), (m).\\
$c_i^\text{travel}$ & Travel capacity of robot $i$, (m).\\
$c_i^\text{work}$ & Work capacity of robot $i$, (m).\\[3pt]
\multicolumn{2}{|l|}{Abbreviations}\\
FACP & Full Area Coverage Path.\\
LWL & Last Worked Location.\\
\hline
\end{tabular}
\end{table}

%


\begin{figure}
\centering
\begin{subfigure}[t]{.5\textwidth}
  \centering
  \includegraphics[width=.7\linewidth]{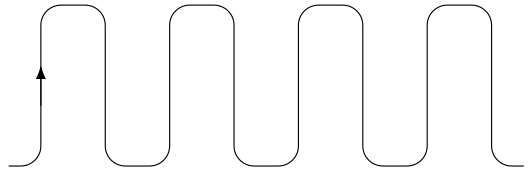}  
\caption{Boustrophedon path planning pattern.\vspace{0.5cm}}
  \label{fig_Boustrophedon_areacovg}
\end{subfigure}
\begin{subfigure}[t]{.5\textwidth}
  \centering
  \includegraphics[width=.8\linewidth]{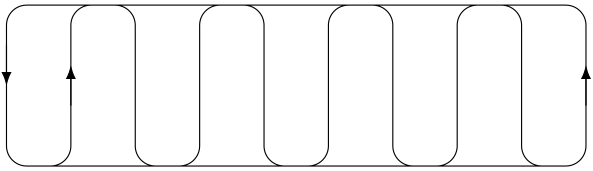}
\caption{Boustrophedon path planning including a headland path.}
  \label{fig_incldheadl_areacovg}
\end{subfigure}
\caption{Visualisation of the Boustrophedon-pattern and and an enclosing headland path.}
\label{fig_boustro}
\end{figure}

\begin{figure}
\centering
\begin{tikzpicture}
\node (image) at (0,0) {\includegraphics[width=6.5cm]{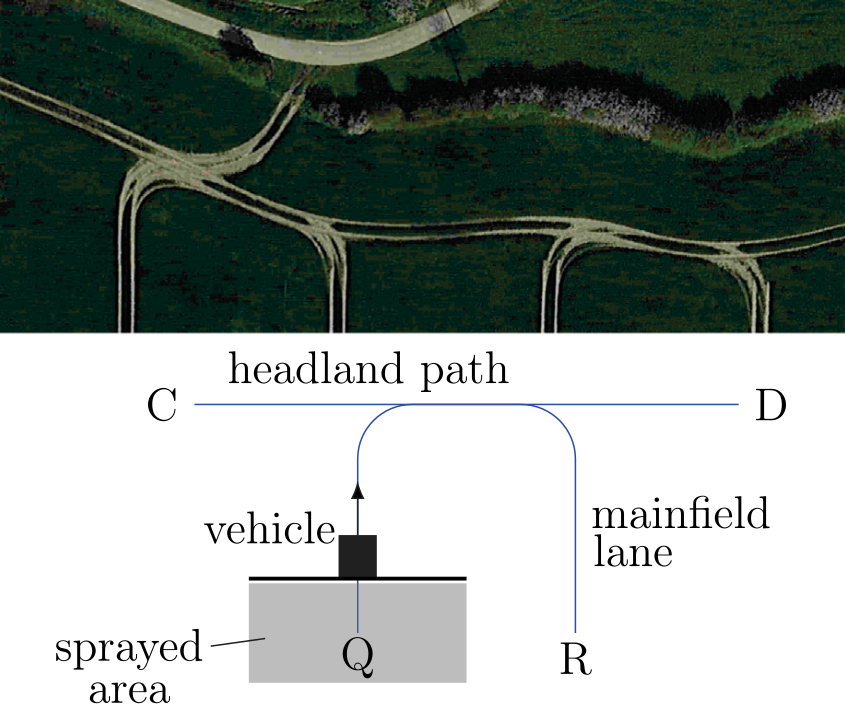}};
\node[color=black,font=\fontsize{9}{11}\selectfont] (a) at (-2.8, -1.4) {boombar};
\draw[black,line width=0.1pt] (-2.15,-1.5) -- (-1.1,-1.72);
\end{tikzpicture}

\caption{Bird's eye view: visualization of the terms \emph{headland path}, \emph{mainfield lane} and a robot covering an area with a working width. One area coverage application is to spray an area (e.g., for a pesticide application). Other agricultural area coverage applications include mowing, fertilizing, seeding, harvesting and so forth. \emph{Top}: Real-world visualization. \emph{Bottom}: Abstract visualization where a robot might travel from location Q towards R or D.}
\label{fig_problFormulation}
\end{figure}

Distributed area coverage promises parallelization of work by distribution of partial area coverage work to different robots or vehicles. Parallelization can lead to reduction of total time required for the area coverage task. In contrast to parallelization, multiple machines can also cooperate to perform sequentially dependent tasks \cite{soitinaho2024heuristic,wang2024multi,mammarella2022cooperation}.

A common objective in distributed area coverage path planning is to minimize the total pathlength collectively traveled by all robots, however other optimization criteria exist also, such as minimizing the number of turns made by robots \cite{vandermeulen2019turn}.

Within an agricultural setting, robots are performing area coverage work typically ground-based. However, the broader topic is likewise relevant for aquatic and aerial applications, for example, for mapping. 

The vast majority of research for distributed area coverage path planning focuses (i) on Boustrophedon path planning-related methods, and (ii) under total absence of a headland path: \cite{kong2006distributed, karapetyan2017efficient, cao2021distributed, agarwal2022area, choton2023optimal, govindaraju2023optimized, gong2024resilient, zheng2025optimal, yakunin2025algorithm, raxit2026energy}. Boustrophedon path planning implies a meandering path pattern as illustrated in Fig. \ref{fig_boustro}. 

Methods that (i) output paths less structured (in comparison to Boustrophedon) or less intuitive coverage paths and which are more closely related to the multiple-Traveling Salesman Problem (m-TSP), but (ii) also lack headland paths consideration include \cite{fazli2013multi, karapetyan2018multi, sun2019novel, huang2020multi, 
zhu2021multi, li2023exact, tang2024efficient, tang2024multi, shen2025multi}. These observations presents both a research gap and the motivation for this paper. A method is presented that explicitly includes a headland path in the distributed area coverage planning problem.

In agriculture, headland paths are common for area coverage applications, especially for cereal crop production. However,  the literature focused so far on single-robot area coverage planning, or dismissed the headland path for the distributed area coverage planning problem.

Headland paths imply a path traversed along the area boundary as visualised in Fig. \ref{fig_problFormulation}. A robot moves along headland path segments and mainfield lanes while conducting area coverage work. An important parameter is the working width of the robot operating in the area, for example, to spray the area with pesticides, fertilizers, and so forth.

Within this paper, the focus is (i) on an alternative path planning method in comparison to the Boustrophedon method, and (ii) the explicit inclusion of a headland path in the distributed area coverage planning problem. In fact, it is argued that both aspects are not a limitation, but, on the contrary, features that improve distributed area coverage path planning.

In this paper, a static and fully observable area is assumed, that can in general be non-convexly shaped and include obstacle areas such as tree islands, ponds and the like. 

The remaining paper is organised as follows: problem formulation, modeling and proposed solution, discussion and the conclusion are described in Sections \ref{sec_materials}-\ref{sec_conclusion}.

\section{Materials and methods\label{sec_materials}}

\subsection{Problem formulation\label{subsec_probformulation}}

The problem addressed can be formulated as follows.

\begin{problem}\label{problem1}
Given an area to be covered that may in general be non-convexly shaped and may include obstacle areas, multiple robots, at least one robot depot, constraints on each robot such as a working width, work capacity limits, travel capacity limits, nonholonomic dynamical constraints, and a possibly given robot start and/or end depot assignment for individual robots, 
calculate  distributed paths for each of the robots such that collectively full area coverage is achieved. 
\end{problem}

\subsection{Discussion of constraints\label{subsec_cstrts}}

\begin{figure}
\centering
\begin{subfigure}[t]{.5\textwidth}
  \centering
  \includegraphics[width=.9\linewidth]{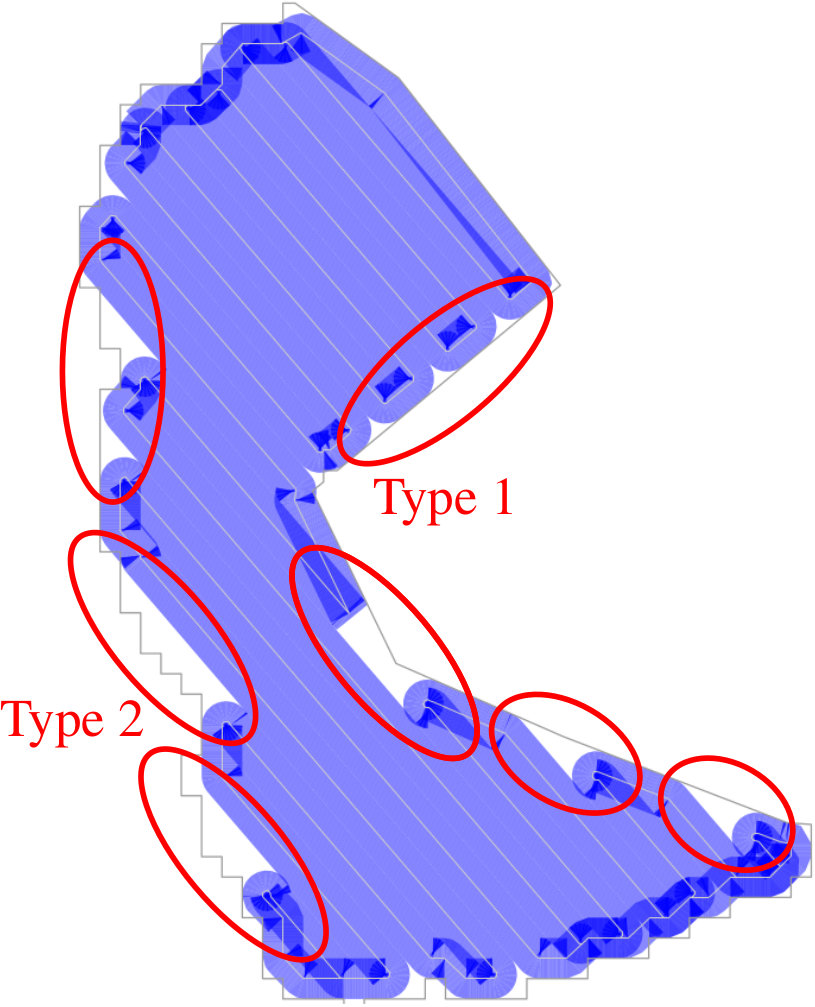}  
%
%
%
\caption{Boustrophedon path: characteristically incurring area coverage gaps. (Type 1) Lanes are circa perpendicular to the area contour. (Type 2) More profound area coverage gaps occur when lanes are almost aligned with the area contour.}
  \label{fig_Boustrophedon_areacovg}
\end{subfigure}
\begin{subfigure}[t]{.5\textwidth}
  \centering
  \includegraphics[width=.8\linewidth]{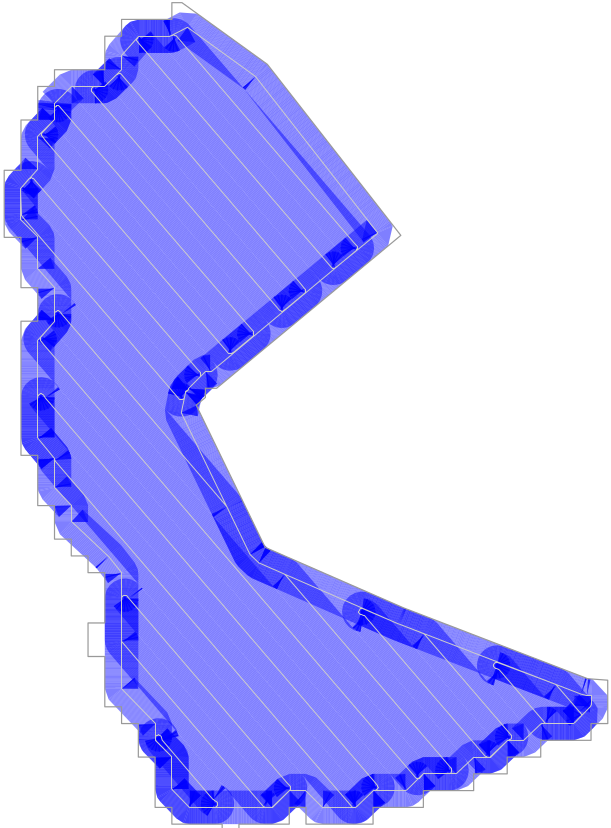}
\caption{Minimised area coverage when also including a headland path.}
  \label{fig_incldheadl_areacovg}
\end{subfigure}
\caption{Illustration of the importance of including a headland path for area coverage applications.}
\label{fig_areacovg}
\end{figure}

The constraints of Problem \ref{problem1} are discussed in detail. Within this paper, a uniform working width $W$ is assumed for all robots. In contrast, work capacity limits $c_i^\text{work}$ (e.g., the spray tank capacity) and travel capacity limits $c_i^\text{travel}$ (e.g., Diesel tank capacity or battery charge) may vary for all robots $i=1,...,N$. 

All robots are further assumed to be subject to nonholonomic dynamical constraints. This implies that for turning all robots require a non-zero turning radius $R_i>0,~\forall i=1,\dots,N$. 

It is assumed $R_i<\frac{W}{2},~\forall i=1,\dots,N$ such that different turning radii for different robots do not impede the general distributed area coverage planning logic. Such an assumption is important since large $R_i$ in combination with transitions for a small angle between mainfield lanes and headland path can otherwise incur area coverage gaps. 

The avoidance of area coverage gaps and overlaps is a main challenge. Area coverage without a headland path incurs area coverage gaps as soon as a robot behaves nonholonomically, which in practice is the case for essentially all robots (at land, air and sea) that operate at velocities larger than very slow. Fig. \ref{fig_areacovg} illustrates the importance of including a headland path for area coverage applications.

Finally, mainfield lanes and a headland path and in particular also the transitions between them may be visible in the area. This is the case for robots with ground-contact, for example, via tyres. The satellite image of Fig. \ref{fig_problFormulation} illustrates this. All areas repressed or compacted by tyre traces are not eligible for crop growth. This motivates to (i) re-use existing tractor traces when covering an area or paths multiple times, and (ii) avoid any new transitions between mainfield lanes and headland path that would create additional compacted areas that can then not be further used for crop growth.


\begin{figure*}
\centering
\begin{subfigure}[t]{.9\textwidth}
  \centering
  \includegraphics[width=.99\linewidth]{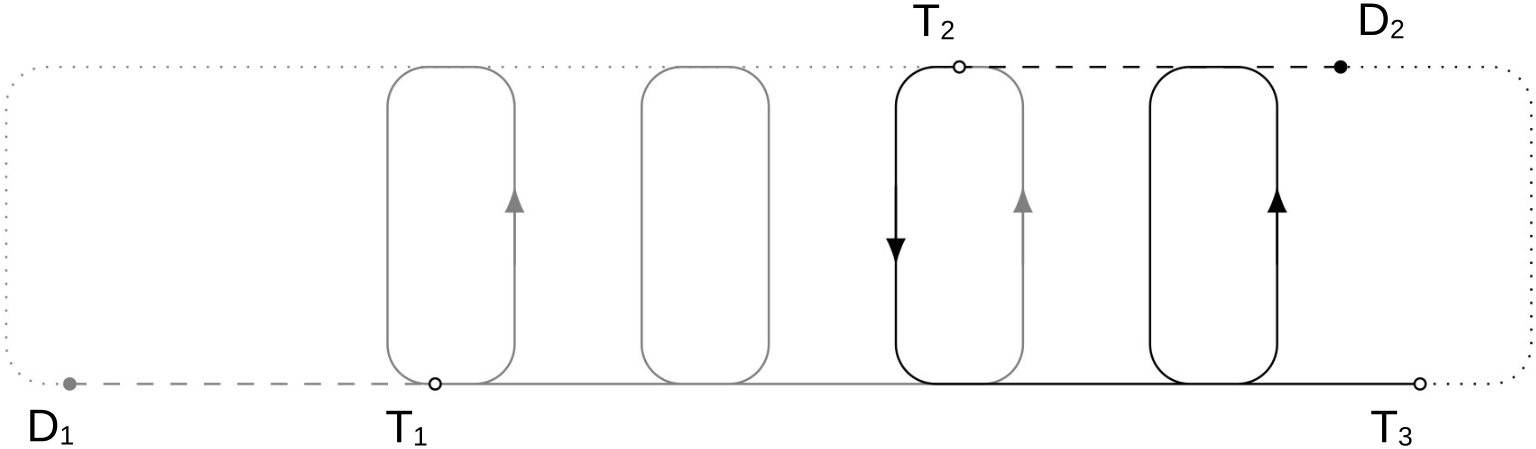}  
\caption{Case 1: $D_i^\text{start}==D_i^\text{exit}$. Each robot starts and ends its travel at the same depot. \vspace{0.3cm}}
  \label{fig_ex1_concept_2robots_Case1}
\end{subfigure}
\begin{subfigure}[t]{.9\textwidth}
  \centering
  \includegraphics[width=.99\linewidth]{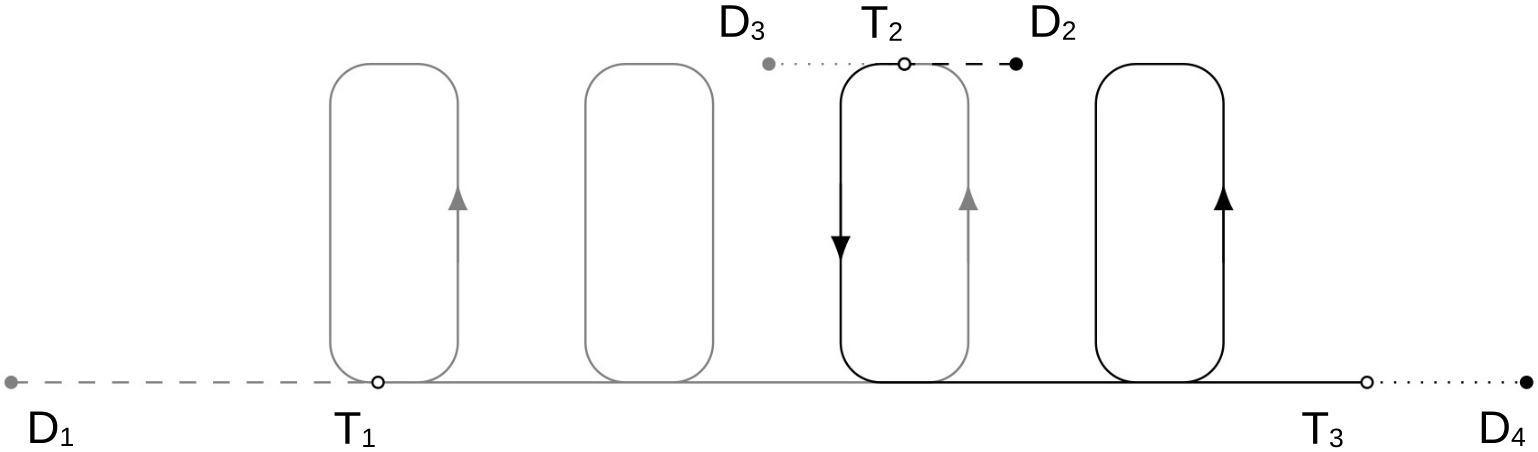}
\caption{Case 2: $D_i^\text{start}\neq D_i^\text{exit}$. The start and exit depot at which a robot starts and ends its travel, respectively, do not need to be equal.}
  \label{fig_ex1_concept_2robots_Case2}
\end{subfigure}
\caption{Illustration of the coordination of two robot paths (in gray and black) for two different cases. Each robot phase consists of (i) a first transition phase (dashed line) that connects the start depot with a path to a location where area coverage begins, (ii) a second phase (solid line) describing area coverage work, and finally (iii) a third phase (dotted line) indicating a path from the last worked location (LWL) to the exit depot. The LWL (location $T_2$) links the two robot paths.}
\label{fig_ex1_concept_2robots}
\end{figure*}

\begin{figure*}
\centering
\includegraphics[width=.7\linewidth]{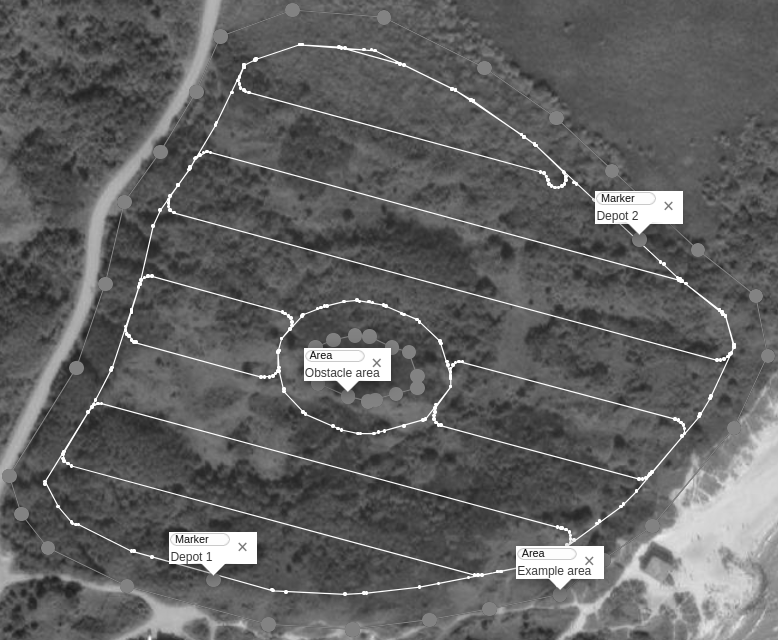}
\caption{Illustration of a real-world example with an obstacle area.}
\label{fig_satfig}
\end{figure*}


\begin{figure*}
\centering
\begin{subfigure}[t]{.8\textwidth}
  \centering
  \includegraphics[width=.99\linewidth]{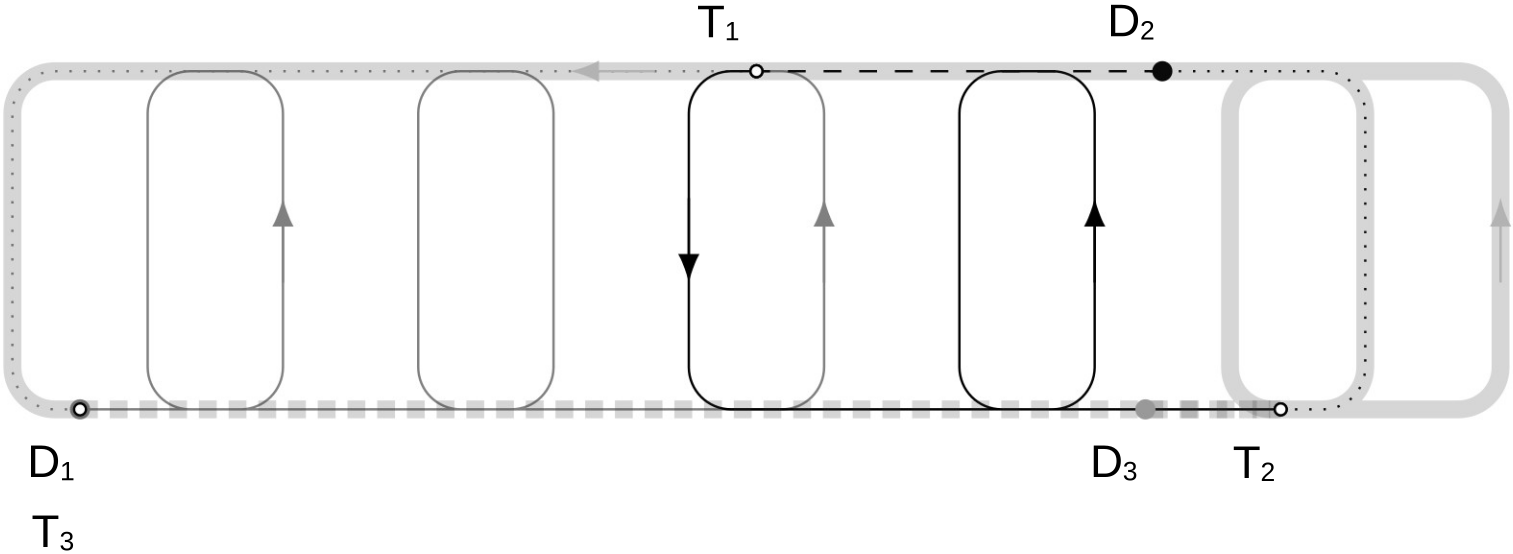}  
\caption{Example 1 with 3 robots and $D_i^\text{start}==D_i^\text{exit},~\forall i=1,\dots,3$. Each robot starts and ends its travel at the same depot. Depot locations are fixed and assumed to be given.\vspace{0.3cm}}
  \label{fig_ex1}
\end{subfigure}
\begin{subfigure}[t]{.8\textwidth}
  \centering
  \includegraphics[width=.99\linewidth]{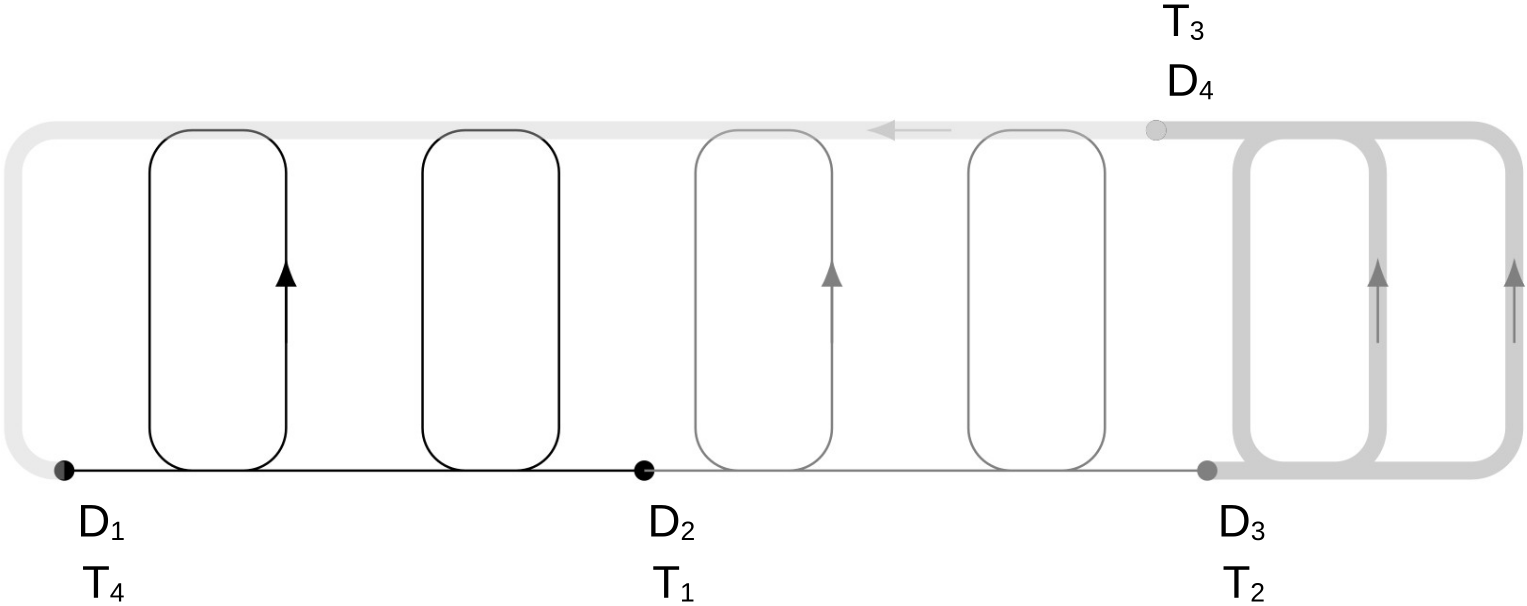}
\caption{Example 2 with 4 robots, $D_i^\text{start}\neq D_i^\text{exit}$, and  free depot placement option. Depots are placed to eliminate all non-working travel of all robots.}
  \label{fig_ex2}
\end{subfigure}
\caption{Two examples to illustrate the effects of (a) fixed depot locations, and (b) optimal free depot placement.}
\label{fig_ex}
\end{figure*}
%
%
%
%

\subsection{Proposed solution\label{sec_solnMethod}}

The proposed distributed area coverage path planning solution is summarised in Algorithm \ref{alg_M1}.

\begin{algorithm}
\SetKwInOut{Subfunctions}{\textbf{Subfunction}}
\SetKwInOut{Input}{\textbf{Data Input}}
\SetKwInOut{Output}{\textbf{Data Output}}
\DontPrintSemicolon
\vspace{0.15cm}
\Subfunctions{$\mathcal{F}^\text{FACP}(\cdot)$,~$\mathcal{F}^{D_1,\text{criterion}}(\cdot)$,~$\mathcal{F}^\text{coverage}(\cdot)$,~$\mathcal{F}^\text{exit}(\cdot)$,~\newline $\mathcal{F}^\text{cost}(\cdot)$,~$\mathcal{F}^{\mathcal{I}}(\cdot)$,~$\mathcal{F}^\text{SP,criterion}(\cdot)$,~$\mathcal{F}^\text{start}(\cdot)$,~$\mathcal{F}^\text{flag}(\cdot)$}
\vspace{0.15cm}\hrule\vspace{0.15cm}
%
\Input{$\mathcal{A},~\mathcal{D}_1,~W,~\{c_i^\text{work},c_i^\text{travel},R_i\}_{i=1}^N$}
%
\vspace{0.15cm}\hrule\vspace{0.15cm}
$\mathcal{S}^\text{opt}, \mathcal{I}^\text{opt}, L_\text{cost}^\text{opt}\leftarrow \{~\},\{~\},\infty$ ~{\color{gray}\%Initialize solution.}\;\vspace{0.15cm}
\For{$D_1 \in \mathcal{D}_1$}
{\vspace{0.15cm}
$P^\text{FACP} \leftarrow \mathcal{F}^\text{FACP}(D_1,W)$~{\color{gray}\%Full Area Coverage Plan.}\;\vspace{0.15cm}
\For{$P \in \{ P^\text{FACP},P_\text{inverted}^\text{FACP} \}$}
{\vspace{0.15cm}
$\mathcal{S},\mathcal{I},L_\text{cost},T_\text{start}^\text{LWL} \leftarrow \{\},\{\}, 0,D_1 $\newline~{\color{gray}\%Initialize candidate solution.}\;\vspace{0.15cm}
$i \leftarrow \arg\max\limits_{i\in\{1,...,N\}}  \mathcal{F}^{D_1,\text{criterion}}(D_1)$, $p_{[D_i, \tilde{T}_\text{start}^\text{LWL}]}^\text{start} \leftarrow  \{\} $\newline~{\color{gray}\%Select a robot according to a criterion.}\;\vspace{0.15cm}
$p_{[T_\text{start}^\text{LWL},T_\text{end}^\text{LWL}]}^\text{coverage},T_\text{end}^\text{LWL} \leftarrow   \mathcal{F}^\text{coverage}(D_1,c_i^\text{work},c_i^\text{travel},P^\text{FACP},\mathcal{D}_i^\text{exit})$\newline~{\color{gray}\%Partial area coverage.}\;\vspace{0.15cm}
$p_{[T_\text{end}^\text{LWL},D_i^\text{exit}]}^\text{exit},D_i^\text{exit} \leftarrow \mathcal{F}^\text{exit}(T_\text{end}^\text{LWL}, c_i^\text{travel},P^\text{FACP},\mathcal{D}_i^\text{exit})$\newline~{\color{gray}\%Shortest path to exit depot.}\;\vspace{0.15cm}
$L_\text{cost}\leftarrow L_\text{cost} + \mathcal{F}^\text{cost}(p_{[D_i, T_\text{start}^\text{LWL}]}^\text{start},~p_{[T_\text{start}^\text{LWL},T_\text{end}^\text{LWL}]}^\text{coverage},~p_{[T_\text{end}^\text{LWL},D_i^\text{exit}]}^\text{exit})$\newline
$\mathcal{S} \leftarrow \mathcal{S} + [p_{[D_i, T_\text{start}^\text{LWL}]}^\text{start},~p_{[D_1,T_\text{end}^\text{LWL}]}^\text{coverage},~p_{[T_\text{end}^\text{LWL},D_i^\text{exit}]}^\text{exit}] $\newline
$\mathcal{I} \leftarrow \mathcal{I} + \mathcal{F}^{\mathcal{I}}(i) $\newline
~{\color{gray}\%Cost calculation and solution update.}\;\vspace{0.15cm}
$i,D_i \leftarrow \arg\min\limits_{i\in\{1,...,N\}\setminus \mathcal{I}}  \mathcal{F}^\text{SP,criterion}(T_\text{end}^\text{LWL}, \mathcal{I}, P^\text{FACP})$\newline~{\color{gray}\%Select a next robot according to a criterion.}\;\vspace{0.15cm}
$p_{[D_i, T_\text{end}^\text{LWL}]}^\text{start} \leftarrow   \mathcal{F}^\text{start}(D_i,T_\text{end}^\text{LWL}, c_i^\text{travel},P^\text{FACP})$\newline~{\color{gray}\%Shortest path to LWL.}\;\vspace{0.15cm}
Follow Steps 7-9.\newline~{\color{gray}\%Partial area coverage plus transition to depot.}\;\vspace{0.15cm}
$\text{flag}^\text{continue} \leftarrow   \mathcal{F}^\text{continue}(\mathcal{S},P^\text{FACP})$\newline~{\color{gray}\%Check if area fully covered.}\;\vspace{0.15cm}
\While{$\text{flag}^\text{continue}==\text{True} $}
{\vspace{0.15cm}
Repeat Steps 10-12.\newline~{\color{gray}\%Continue assignment of robots.}\;\vspace{0.15cm}
$\text{flag}^\text{continue} \leftarrow   \mathcal{F}^\text{continue}(\mathcal{S},P^\text{FACP})$\newline~{\color{gray}\%Check if area fully covered.}\;\vspace{0.15cm}
} 
\If{$L_\text{cost}<L_\text{cost}^\text{opt} $}
{\vspace{0.15cm}
$\mathcal{S}^\text{opt},\mathcal{I}^\text{opt},L_\text{cost}^\text{opt} \leftarrow   \mathcal{S},\mathcal{I},L_\text{cost}$\newline~{\color{gray}\%Update current best solution.}\;\vspace{0.15cm}
} 
} 
} 
%
\vspace{0.0cm}\hrule\vspace{0.15cm}
\Output{$ \mathcal{S}^\text{opt},\mathcal{I}^\text{opt},L_\text{cost}^\text{opt} $~{\color{gray}\%Individual path plans for a set of robots that collectively achieve area coverage.}}
\vspace{0.05cm}
\caption{Distributed area coverage path planning subject to the inclusion of a headland path}\label{alg_M1}
\end{algorithm}

Several comments can be made. Each individual robot is labeled by index $i\in\{1,\dots,N\}$. 

Each robot is initially located at a \emph{start depot} from where it will begin its partial area coverage work. Depending on a given application, this assignment can  (i) be assumed as being given, or (ii) can be assigned by the method.

The notion of a \emph{principal depot} is introduced. The principal depot denotes a depot, which is in general not different from other depots, but which plays an important role  for the calculation of a full area coverage path plan (FACP) in Step 3 of Alg. \ref{alg_M1}. FACP starts at the principal depot. Here, the method from \cite{plessen2019optimal} is used for the calculation of FACP.

$\mathcal{D}_1$ describes the set of all depots which are admissible to be selected as principal depot and at which there is at least one robot initially located. For a typical application, $\mathcal{D}_1$ would describe the set of all depots.

In Step 4 of Alg. \ref{alg_M1} both the FACP plan,  $P^\text{FACP}$, and its inverse (the path given by the inversion of the sequence of position coordinates), here denotes as $P_\text{inverse}^\text{FACP}$ , are considered. Note that this inversion does not change the the FACP path length.

Then, remaining Steps 5-18 are based on a specific FACP plan candidate.

In Step 6 at the principal depot candidate $D_1$, a robot $i$ is selected according to a criterion. Depending on the problem constraints, a robot may, for example, be selected as (i) the one with maximal work capacity and minimial activation cost in case there are multiple robots at $D_1$, or (ii) the one with maximal work capacity and minimial activation cost among all available robots in case of a free assignment of robots to depots is permitted.

In Step 7 the selected robot conducts partial area coverage work (for example, spraying work) as long as its work capacity $c_i^\text{work}$ this permits, and its travel capacity $c_i^\text{travel}$ (i.e., driving without conducting spraying work) is such that it still permits a transition along the shortest path \cite{plessen2016shortest} along the transition graph towards a designated robot exit depot $D_i^\text{exit}$ . The transition location where coverage work is stopped shall be abbreviated as the \emph{last worked location} $T_\text{end}^\text{LWL}$. 

A first option to select $T_\text{end}^\text{LWL}$ as the location where the work capacity is full exhausted, i.e., where $c_i^\text{work}=0$. However, a second option is to select a location slighly before that along the partial area coverage path and where the capacity is almost but not yet fully exhausted with $c_i^\text{work}=\epsilon>0$ for a small $\epsilon$. For this option $T_\text{end}^\text{LWL}$ is selected as the last possible location along an Eulerian edge along the transition graph and according to the FACP where $c_i^\text{work}$ is not yet fully exhausted and travel capacity $c_i^\text{travel}$ is such that it still permits a transition along the shortest path along the transition graph towards a designated robot exit depot $D_i^\text{exit}$.

This second option offers the advantage that the shortest path from $T_\text{end}^\text{LWL}$ towards a designated exit depot $D_i^\text{exit}$ is in general shorter. For illustration of the location selection for $T_\text{end}^\text{LWL}$, see Fig. \ref{fig_ex1_concept_2robots} where exemplarily $T_\text{end}^\text{LWL}=T_2$. The transition location $T_\text{end}^\text{LWL}$ will later link partial area coverage paths between different robots in Step 11, as will be further detailed below.

Step 8 works in tandem with Step 7. The shortest path from $T_\text{end}^\text{LWL}$ towards an exit depot $D_i^\text{exit}$ is calculated. It can be distinguished between different scenarios depending on the problem setup. An exit depot $D_i^\text{exit}$ may (i) be fixedly assigned to robot $i$, or (ii) may be freely selected out of a set of available exit depots $\mathcal{D}_i^\text{exit}$, whereby in all cases remaining travel capacity $c_i^\text{travel}$ must be respected for the shortest path transition.

Step 9 summarises the cost incurred by operation of robot $i$, which may include travel cost, area coverage work cost, and activation cost of start and exit depot. It further stores the path taken by the robot. Finally, depending on problem setup, robot $i$ is (i) labelled as a robot not further eligibible for the remainder of the distributed area coverage task, or (ii) labelled as a robot stationed at $D_i^\text{exit}$ but still eligible for the remainder of the distributed area coverage task under the assumption that the robot's work and travel capacity can there be recharged.

In Step 10, a next robot candidate for partial area coverage work is selected according to a criterion. This is typically done by selecting a robot that offers the shortest path from its depot towards $T_\text{end}^\text{LWL}$. In case there are multiple robot candidates offering the shortest path, a robot may, for example, be selected as (i) the one with maximal work capacity and minimial activation cost, or (ii) the one with maximal work capacity and minimial activation cost among all available robots in case of a free assignment of robots to depots is permitted.

Step 11 summarizes the shortest path from robot start depot towards $T_\text{end}^\text{LWL}$. At the latter location partial area coverage begins for the robot according to part of the FCAP-plan. Accordingly Step 7-9 are repeated for the robot.

Thus, $T_\text{end}^\text{LWL}$ represents the transition locations at which partial area coverage work of two robots is synchronized. See Fig. \ref{fig_ex1_concept_2robots} for illustration. It can be further differentiated between two cases. First, if a start and exit depot for a robot are equal, a scenario as illustrated in Fig. \ref{fig_ex1_concept_2robots_Case1} results. Second, if start and exit depot for a robot do not necessarily have to be equal, a scenario as illustrated in Fig. \ref{fig_ex1_concept_2robots_Case2} results.

Then, Steps 13-18 of Alg. \ref{alg_M1} indicate how above steps are repeated to assign partial area coverage paths to different robots until the given area is collectively fully covered.

In Step 18 the best solution out of the solution candidates according to Step 2 and 4 are selected.

Fig. \ref{fig_satfig} illustrates a result when also obstacle areas are considered.

Finally, an important application aspect is stressed. While partial area coverage paths are planned according to Alg. \ref{alg_M1} in sequence to coordinate all robots, during application the work of all robots can be conducted fully in parallel with all robots operating simultaneously. This is enabled by the fact that all partial paths are coordinated via the LWL along the FACP-plan.


%


\begin{figure}
\centering
\begin{subfigure}[t]{.5\textwidth}
  \centering
  \includegraphics[width=.5\linewidth]{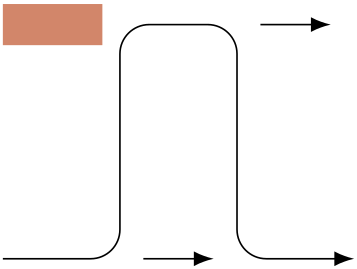}  
\caption{A Boustrophedon pattern. The red bar indicates the area that cannot be reached by neither traversal of the Boustrophedon-pattern nor traversal of the headland segments in the directions as indicated by the arrows.}
  \label{fig_Boustrophedon_areacovg}
\end{subfigure}
\begin{subfigure}[t]{.5\textwidth}
  \centering
  \includegraphics[width=.8\linewidth]{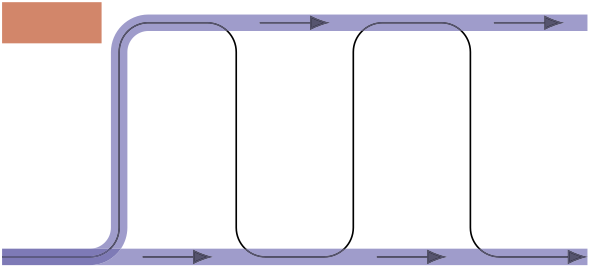}
\caption{Concatenation of two Boustrophedon-patterns. The traversal along the "upper" and "lower" headland path is emphasised in blue. Importantly, the area indicated by the red bar can still not be reached.}
  \label{fig_incldheadl_areacovg}
\end{subfigure}
\caption{Counterexample 1: The repressed-area problematic for Boustrophedon-based path planning. See Sect. \ref{subsec_counterex} for a detailed discussion.}
\label{fig_counterex1}
\end{figure}


\begin{figure*}
\centering
\includegraphics[width=.99\linewidth]{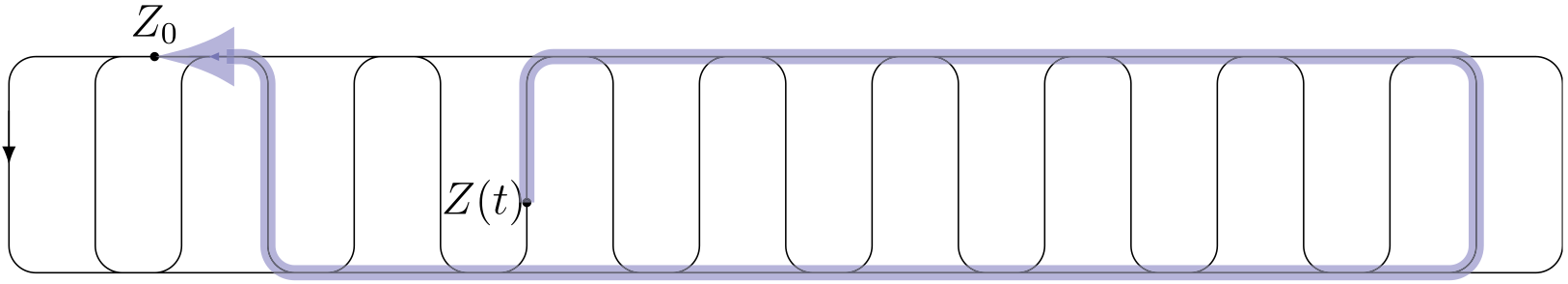}
\caption{Counterexample 2: The repressed-area problematic for Boustrophedon-based path planning. It a robot located at $Z(t)$ wishes to reach transition $Z_0$ subject to the repressed area constraint, it can do so only via a large detour. See Sect. \ref{subsec_counterex} for a detailed discussion.}
\label{fig_counterex2}
\end{figure*}


%

\section{Discussion and limitations\label{sec_IllustrativeEx}}

\subsection{Unconstrained depot assignment and global optimality}

Fig. \ref{fig_ex} illustrates two examples to highlight the difference when depot locations are fixed or can be freely assigned. 

In the latter case, a global optimal solution for the distributed area coverage path planning problem can be constructed. This can be seen as follows. By non-negativity of pathlengths a global optimal FACP solution exists. It can be partitioned into partial paths, whereby each partial path is served by a corresponding robot. The start and end point of each partial paths are then assigned as the depot locations, whereby each depot serves as the start depot to one robot and as the exit depot of another robot. It remains that each robot starts each partial area coverage path at its designated start depot.

Two comments can be made. First, all partial area coverage paths can be worked in parallel by the robots, which is the key advantage over a single robot working the FACP. The disadvantage is that robot depots (an infrastructure) is needed. 

Second, the free depot assignment permits to eliminate initial and final non-working transition phases for all robots. This in consequence enables the optimality of the solution.

Finally, an even stronger scenario can be constructed. If additionally it is assumed that (i) all depots permit to recharge (both work as well as travel capacacity) at all depots, and (ii) each partial path is selected such that it can be worked by even the robot with least work capacity, then area coverage in continuum is possible, where a continuous work pattern of robots can be achieved. This implies that each robot traverses each depot and is recharged at each depot.

\subsection{Discussion of heuristic Step 6 and 10}

Step 6 and 10 of Alg. \ref{alg_M1} imply a heuristic selection step. If there are multiple robots at a depot one robot has to be selected according a criterion. Such a criterion may be, for example, to prioritize a robot with maximal work capacity, $c_i^\text{work}$, out of the options at the depot. This heuristic may be sensible, but it is noted that it does not guarantee to yield a global optimal solution to Problem \ref{problem1}. 

In contrast, the global optimal solution (which exists by non-negativity of path lengths) can be achieved by exhaustively testing all robot selection options and calculating all corresponding downstream solution dependencies, thereby implicitly constructing a solution tree. Unfortunately, the size of this tree grows exponentially with the number of robots involved. Therefore a heuristic is more appropriate.

\subsection{Collision avoidance\label{subsec_collisionavoidance}}

Since all robots are coordinated via the FACP-plan in Step 2 of Alg. \ref{alg_M1}, frontal collision scenarios of robots are avoided by construction. Because of the FACP-coordination all robots are always driving in the same direction along headland path edges. Therefore, any heuristic priority scheme (e.g., the robot closest to a shared headland path segment has priority to pass that segment first) is sufficient for local collision avoidance.

\subsection{Discussion of counterexamples for alternative methods\label{subsec_counterex}}

Several counterexamples and arguments are provided to motivate why (i) Boustrophedon-based planning, and (ii) the absence of a headland path is not a viable alternative to the presented method.

First, in case of absence of a headland path coverage possibly very large area coverage gaps can result. This is illustrated in Fig. \ref{fig_areacovg}. Therefore, using only a Boustrophedon pattern for area coverage is suboptimal. This statement holds for both, single-robot area coverage and distributed multi-robot area coverage.

Second and following up on the previous argument, if a headland path coverage is sought to be included, then the combination of Boustrohpedon-pattern with headland path coverage is suboptimal as demonstrated in \cite{plessen2018partial, plessen2019optimal, plessen2025predictive}.  

Third, the Boustrophedon pattern in combination with shortest path transitions, in particular, within agricultural fields  are problematic. This was discussed in \cite{plessen2018partial} and is further illustrated by Fig. \ref{fig_problFormulation}, \ref{fig_counterex1} and \ref{fig_counterex2}. For example, for robots with ground-contact tyre-traces in the ground repress or compact area that is then not further eligible for crop growth (cf. \ref{fig_problFormulation}). This motivates to (i) re-use existing tractor traces
when covering an area or paths multiple times, and (ii) avoid
any new transitions between mainfield lanes and headland path
that would create additional compacted areas that can the no longer be used for crop growth. For distributed area coverage multiple shortest paths between depots and LWL-locations are needed. Subject to the constraint of re-using only existing tractor traces for movement and avoiding new repressed area transitions, scenarios such as illustrated in Fig. \ref{fig_counterex2} typically result, which imply very long detours and path lengths accordingly.

Fourth, while the previous first two argument remain valid the last third argument about repressed area transitions is not relevant for aquatic or aerial applications. In such applications the combination of Boustrophedon-pattern with headland path coverage for distributed area coverage \emph{may} be shorter than the proposed solution in Sect. \ref{sec_solnMethod}. However, no generalizing statement can be made (examples can be constructed or selected in favor of one or another solution). Shortest path contributions to the overall path length now become shorter because the transition graph is less constrained, but the overall path length may still be larger since the combination of Boustrophedon-pattern and headland path coverage is suboptimal for area coverage as shown in \cite{plessen2018partial, plessen2019optimal}. This suboptimality may, in general, outweight the savings from the shorter path contributions. 

Note that Alg. \ref{alg_M1} can likewise be used for distributed area coverage path planning based on the Boustrophedon-pattern with headland path coverage. Therefore, only Step 3 in Alg. \ref{alg_M1} needs to be adapted to account for Boustrophedon-based planning. 

Similarly, note that freeform mainfield lanes \cite{plessen2021freeform} or 3D data \cite{plessen20262d,bostelmann2023multi} only changed the transition graph coordinates, but do not require a change of the structure of Alg. \ref{alg_M1}.

\section{Conclusion\label{sec_conclusion}}

A deterministic algorithm for distributed area coverage path planning subject to the inclusion of a headland path was presented. Various variations were discussed. A scenario was presented where unconstrained depot locations assignment along the headland path can attain a global optimal solution. 




The importance of inclusion of a headland path for area coverage, as well as the benefits in comparison to an alternative Boustrophedon-pattern based approach were highlighted.

\bibliography{mybibfile.bib}
\nocite{*}







\end{document}